\documentclass[]{llncs}

\usepackage[T1]{fontenc}
\usepackage{verbatim}
\usepackage{lmodern}
\usepackage{amssymb,amsmath}
\usepackage{ifxetex,ifluatex}
\usepackage{fixltx2e} 
\usepackage{csvsimple}
\usepackage[title]{appendix}
\usepackage{longtable,booktabs}
\usepackage{makeidx}
\institute{Ano  \\ \texttt{}}

\IfFileExists{upquote.sty}{\usepackage{upquote}}{}
\ifnum 0\ifxetex 1\fi\ifluatex 1\fi=0 
  \usepackage[utf8]{inputenc}
\else 
  \ifxetex
    \usepackage{mathspec}
    \usepackage{xltxtra,xunicode}
  \else
    \usepackage{fontspec}
  \fi
  \defaultfontfeatures{Mapping=tex-text,Scale=MatchLowercase}
  
\fi
\IfFileExists{microtype.sty}{\usepackage{microtype}}{}
\usepackage{graphicx}
\makeatletter
\def\ScaleIfNeeded{%
  \ifdim\Gin@nat@width>\linewidth
    \linewidth
  \else
    \Gin@nat@width
  \fi
}
\makeatother
\let\Oldincludegraphics\includegraphics
{%
 \catcode`\@=11\relax%
 \gdef\includegraphics{\@ifnextchar[{\Oldincludegraphics}{\Oldincludegraphics[width=\ScaleIfNeeded]}}%
}%
\ifxetex
  \usepackage[setpagesize=false, 
              unicode=false, 
              xetex]{hyperref}
\else
  \usepackage[unicode=true]{hyperref}
\fi
\hypersetup{breaklinks=true,
            bookmarks=true,
            pdfauthor={Anon},
            pdftitle={Learning to segment neuroanatomy from small numbers of weakly labeled cases},
            colorlinks=true,
            citecolor=blue,
            urlcolor=blue,
            linkcolor=magenta,
            pdfborder={0 0 0}}
\title{Few-shot brain segmentation from weakly labeled data with deep heteroscedastic multi-task networks}
\author{Richard McKinley\inst{1} \and Michael Rebsamen\inst{1} \and Raphael Meier\inst{1} \and Mauricio Reyes\inst{2} \and Christian Rummel\inst{1} \and Roland Wiest\inst{1}}
\date{}
\institute{Support Centre for Advanced Neuroimaging, University Institute of Diagnostic and Interventional Neuroradiology, Inselspital, Bern University Hospital, Bern, Switzerland 
\and
Insel Data Science Centre, Inselspital, Bern University Hospital, University of Bern, Switzerland}

\begin{document}
\maketitle
\begin{abstract}
In  applications of supervised learning applied to medical image segmentation, the need for large amounts of labeled data typically goes unquestioned.  In particular, in the case of brain anatomy segmentation, hundreds or thousands of weakly-labeled volumes are often used as training data. In this paper, we first observe that for many brain structures, a small number of training examples, (n=9), weakly labeled using Freesurfer 6.0, plus simple data augmentation, suffice as training data to achieve high performance, achieving an overall mean Dice coefficient of $0.84 \pm 0.12$ compared to Freesurfer over 28 brain structures in T1-weighted images of $\approx 4000$ 9-10 year-olds from the Adolescent Brain Cognitive Development study.  We then examine two varieties of heteroscedastic network as a method for improving classification results. An existing proposal by Kendall and Gal, which uses Monte-Carlo inference  to learn to predict the variance of each prediction,  yields an overall mean Dice of $0.85 \pm 0.14$ and showed statistically significant improvements over 25 brain structures. Meanwhile a novel heteroscedastic network which directly learns  the probability that an example has been mislabeled yielded an overall mean Dice of $0.87 \pm 0.11$ and showed statistically significant improvements over all but one of the brain structures considered.  The loss function associated to this network can  be interpreted as performing a form of learned label smoothing, where labels are only smoothed where they are judged to be uncertain.


\end{abstract}

\section{Introduction}

Manual segmentation of volumetric medical data, such as magnetic resonance imaging, is a laborious, time-consuming task, with very high inter-rater variability. This limitation means that high quality labeled data for the training and validation of machine-learning methods can only be found in small quantities, and that larger labeled datasets typically contain a large amount of label noise. As a result, it is of utmost importance for the field that robust methods be found to learn from small amounts of data, from noisy labeling of data, or in the worst case from small amounts of noisy data. For the majority of medical imaging tasks, there is no reasonable alternative to training from at least some manually labeled data.  However, for the segmentation of normal-appearing brain structures there are a number of freely available non-learning-based tools which can approximate a manual segmentation. This, coupled with the availability of increasing large research datasets of disease populations and healthy controls, has led to many researchers training models solely or partially on hundreds or thousands of scans together with ``auxiliary labels'': automated segmentations derived from existing tools.\cite{rajchl2018,GUHAROY2019,Mcclure2018}   In each case  the model performance of the trained method was attributed to the size of the training set.  However, in other fields of computer vision, segmentation problems are tackled with dramatically smaller amounts of data.  The CamVid street scene segmentation problem, for example, provides 367 (2D) segmented images for training, and tackles a much more heterogeneous problem than brain segmentation.  Given the strong spatial priors inherent in the task, it would be surprising if brain image segmentation required more data than natural image segmentation.  In this paper, we examine the feasibility of learning brain segmentation from a very small number of cases, and only from auxiliary labels.  Our goals are i) to assess the feasibility of learning in such a small data domain, and ii) to investigate the benefit of estimating aleatoric uncertainty via heteroscedastic classification. Heteroscedastic classification networks which predict the variance of their outputs were introduced in \cite{Kendall2017WhatUD}, where it was shown to improve street-scene segmentation: this increase in performance can be attributed to learned loss attenuation, in which gradients from examples with possibly erroneous labels are attenuated.  The use of heteroscedastic classification networks in medical image segmentation has been so far limited, with authors focusing on uncertainty derived from dropout \cite{Mcclure2018,Jungo2018,Jungo2018b,Roy2018} or test-time augmentation\cite{WANG2019}. Predictive variance was explored, together with other measures of uncertainty, as a method of filtering MS lesion segmentations by Nair et al~\cite{Nair2018}.  A multi-task network using a homoscedastic (per task rather than per example) measure of task uncertainty was presented by Bentaib et al~\cite{Bentaieb2017}.  A direct application of predictive variance as formulated for regression was applied to classification in Bragman et al \cite{Bragman2018}, but the effect on segmentation performance was not assessed.    In this paper, we focus on the benefit of two variants of heteroscedastic networks: predictive variance, and a new "label-flip" uncertainty measure.  In this second method the network predicts, for each voxel and task, the probability that the model output will disagree with the ground truth: these probabilies are then used to perform learned label smoothing \cite{Szegedy2016}.  We train and test our method on data from the ABCD cohort of 9-10 year olds.\cite{VOLKOW2018,CASEY2018}  Since we utilize only a small amount of data for training and validating our model, we can compare methods on an extremely large number of  test cases ($n \approx 4000$).

\section{Label uncertainty and heteroscedastic classification}

\begin{figure}
    \centering
    \includegraphics[width=10cm, keepaspectratio]{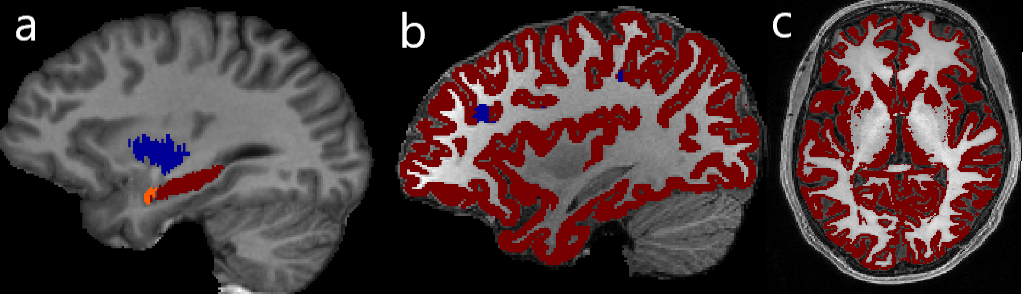}
    \caption{Three examples of ground truth error: a) random, predictable errors in the Putamen (blue), Amygdala (orange) and Hippocampus (red) in a manually labeled case from the Multi-Atlas-labeling challenge, caused by labeling only in the coronal plane b)  random, sporadic error by Freesurfer: tissue is erroneously labeled as white-matter hypointensity (blue) rather than grey matter (red) in an unpredictable location c) systematic under segmentation of grey matter in the Putamen and Thalamus by FSL-FAST.}
    \label{fig:errors}
\end{figure}

By an error in a segmentation, we mean a disagreement between two label sets, whether they are manually generated, auxiliary labels, or the output of a deep-learning model.
We distinguish in this paper between two categories of ground-truth error.  Systematic errors are those made consistently by a rater (either a human rater or an automated method) across most cases. Learning a correct segmentation from a ground truth containing systematic errors is therefore essentially impossible, as the classifier will learn to make the same errors as in the ground truth. Random errors essentially refer to label noise: with some probability, the label assigned will be incorrect.  We refer to two different kinds of random errors: "predictable" random errors are those where a learner (human or machine) can learn to predict where an error might occur, and with what frequency, while 'sporadic' random errors either occur completely at random, or are so rare that their occurrence and probability cannot be estimated.  For examples of each of these error types see Figure~\ref{fig:errors}.

To make the distinction between predictable and sporadic errors concrete, we need a classifier which can learn where labeling errors are likely.  The term ``heteroscedastic regression'' refers to regression models which do not assume constant variance of residuals, but rather predict both the mean and the variance of the predicted quantity.\cite{Nix94} This notion of uncertainty is distinct from Bayesian Uncertainty, as approximated using Bayesian Dropout techniques; the two were contrasted and presented in a combined form by Kendall and Gal \cite{Kendall2017WhatUD}.  For classification problems, the correct notion of heteroscedasticity is not immediately clear.  We consider two different formulations of heteroscedastic classifier.

\textbf{Predictive variance}
In the predictive variance method of Kendall and Gal \cite{Kendall2017WhatUD}, as applied to binary classification, the logit outputs of the network (i.e. the output of the network before application of a sigmoid nonlinearity) are assumed to follow a Gaussian distribution.  For each example the network outputs a probability, $p$, and a log variance, $\log(\sigma^2)$.  Unlike for heteroscedastic regression the loss function cannot be computed as an analytic function of $p$, $\sigma^2$ and $x$, the true label.  Instead the loss is approximated by averaging a loss not involving $\sigma^2$ over $T$ Monte-Carlo samples, in each of which the logit is perturbed by a normally distributed noise term with mean zero and s.d. $\sigma$.

\textbf{Heteroscedastic Label-flip uncertainty}
In this paper we make use of a new kind of heteroscedastic network, in which the uncertainty in the logit is directly modelled by the probability of disagreement with the ground truth, or 'label-flip'.  For each example of a binary classification problem, the network outputs a $ p \in (0,1)$ denoting class membership, and an output $q \in (0, 0.5)$ predicting the probability that the ground truth and classifier disagree.  If $x \in \{ 0, 1 \}$ is the label of the voxel, according to the (weak) ground truth, and L is a loss function (for example, binary cross-entropy or focal loss), the \emph{label-flip loss} at to that voxel is

\begin{equation}
L(p, (1-x)*q + x*(1-q)) + L(q, z)
\label{eq:fliploss}
\end{equation}

where 
z
is the indicator function for disagreement between the classifier (thresholded at the $p=0.5$ level) and the (weak) ground truth.  Unlike for predictive variance this loss can be formulated in closed form and is differentiable, and so can be used directly in backpropogation. Label-flip loss can be seen as a form of loss attenuation: the loss at voxels with low label noise is dominated by the first loss term, and the loss at voxels with substantial label noise is dominated by the second term.  It can also be seen as learned \emph{label smoothing}: a hard labels are replaced by a soft labels according to the uncertainty in the data.~\cite{Szegedy2016}.  

 Sporadic errors in the training set may now be defined as examples where a) the trained model disagrees with the ground truth, and b) the uncertainty at this example is very low. We propose therefore general scheme of filtering training examples: gradients from examples which disagree with the ground truth, but are classified with with high certainty, are masked during training.

\section{Experimental Setup}
\subsubsection{Data and Labels}
We tested the hypotheses that a) brain segmentation is feasible from small amounts of weakly labeled data, and b) that heteroscedastic networks and sporadic error rejection can improve segmentation results in few shot learning from weak labels, by training both plain and heteroscedastic networks on segmentations of T1 weighted imaging data from the early release of the ABCD dataset\cite{VOLKOW2018,CASEY2018}.  These data were collected from 9 and 10 year olds scanned at 21 different sites, using scanners from three different vendors. Labels for the training, validation and test cases were generated using Freesurfer 6.0. The T1w volumes were skull-stripped using a freely-available tool\footnote{\url{https://github.com/placeholder/github/repo/for/anonymity}}, and the non-zero voxels in each volume were rescaled to have mean zero and unit standard deviation.
We segment a subset of the labels segmented by Freesurfer: Cortical White Matter (L/R), Cortical Grey Matter (L/R), Lateral Ventricles (L/R), Cerebellum (L/R), Thalamus (L/R), Caudate (L/R), Putamen (L/R), Pallidum (L/R), Accumbens Area (L/R), Hippocampus (L/R), Amygdala (L/R), Ventral DC (L/R), 3rd ventricle, 4th ventricle, Brainstem and Corpus Callosum. By contrast with QuickNAT~\cite{GUHAROY2019}, we segment the Accumbens area: in addition, we do not distinguish between cerebellar grey and white matter, and we include the inferior lateral ventricles in the lateral ventricle labels.  As an auxilliary task, we predict labels for the total left hemisphere, the total right hemisphere, and brain parenchyma (voxels belonging to any brain structure, plus WM hypointensities).  
In our experiments we selected three cases from each vendor for training (leading to a total of nine training examples) and two from each vendor for validation during training.  This means that a substantial amount of data remains for testing our methods: we applied our trained classifiers to 4069 additional cases. 

\subsubsection{Model and training}
\begin{figure}
    \centering
    \includegraphics[width=12cm, keepaspectratio]{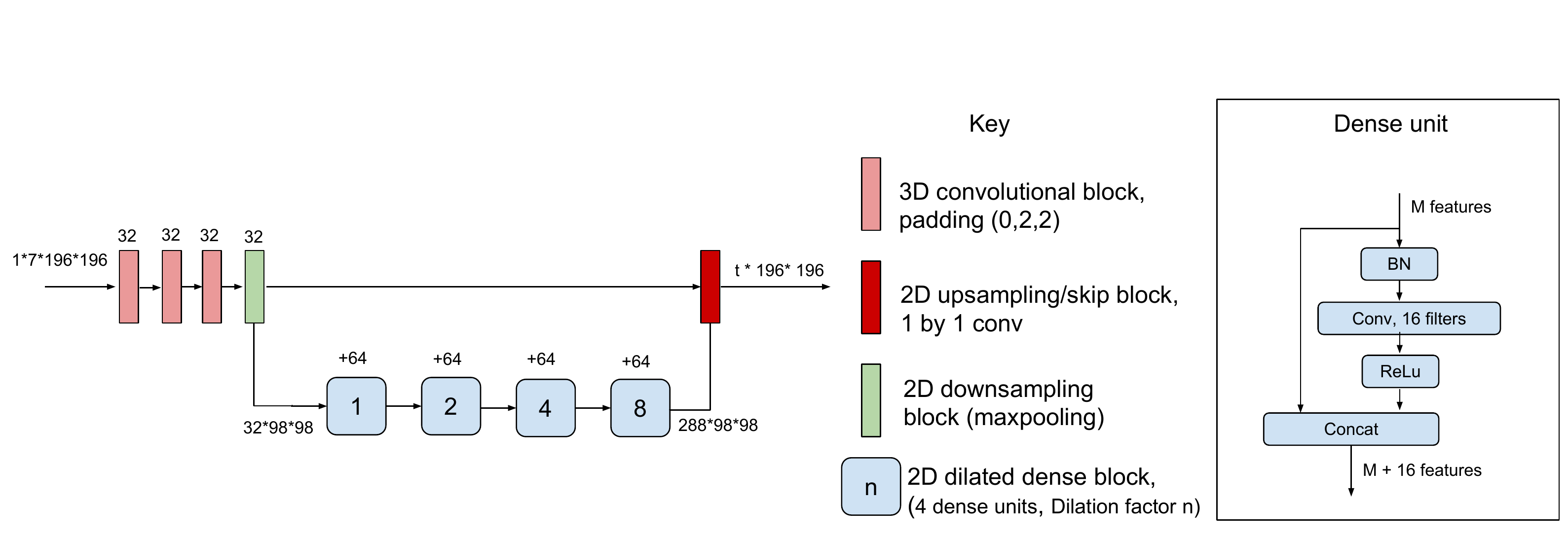}
        \caption {The model architecture used for all three networks considered.  The parameter $t$ is the number of logit outputs (31 for the plain model, 62 for the heteroscedastic models).}
    \label{fig:deepscan}
\end{figure}
Our model architecture (shown in Figure \ref{fig:deepscan}) was implemented in Pytorch: it consists of an initial phase of 3D convolutions to reduce a non-isotropic 3D patch to 2D, followed by a swallow encoder/decoder network using densely connected dilated convolutions in the bottleneck. We use multi-task rather than multi-class classification: each brain region is treated as a separate binary classification problem.  This enables us to use the simplified formulation of heteroscedastic networks as presented above, but is also appropriate in medical imaging, where partial volume effects mean that it is not inconsistent to assign more than one tissue label to a voxel.  For some tissue classes considered, the class imbalance between tissue and background is as high as 2500 to one.  To combat this problem, we use focal loss \cite{Lin2017FocalLF} with parameter $\gamma=2$.  The loss functions for our heteroscedastic networks use focal loss as a base loss function.   Inputs to the network (7*196*196 patches) were sampled randomly from either axial, sagittal or coronal direction.  We perform simple data augmentation: reflection about the (approximate) midline, rotation around a random principal axis through a random angle, and global shifting/rescaling of voxel intensities.  The network was trained with RMSprop, using a batch size of 2 and a cosine annealing learning rate schedule with restarts~\cite{Loshchilov2016}, where the learning rate was varied from $10^{-4}$ to $10^{-7}$ every 2000 steps.   We trained one network for 520 restarts with no uncertainty loss, one model for 20 restarts with no uncertainty loss and then for 500 restarts with predictive variance loss, and one network for 20 restarts with no uncertainty loss and then for 100 restarts with label-flip loss, followed by 400 restarts with label-flip loss and sporadic error rejection: every 20 restarts, the classifier was run over the training set, and voxels which disagreed with the ground truth in all three of the sagittal, coronal and axial views with low flip probability ($<0.001$) were masked from training for the next 20 epochs. Final segmentations were derived by ensembling axial, sagittal and coronal views by averaging logits, and were compared using the Wilcoxon signed rank test.
\section{Results}
\begin{figure}
 \centering
    \includegraphics[width=12cm, keepaspectratio]{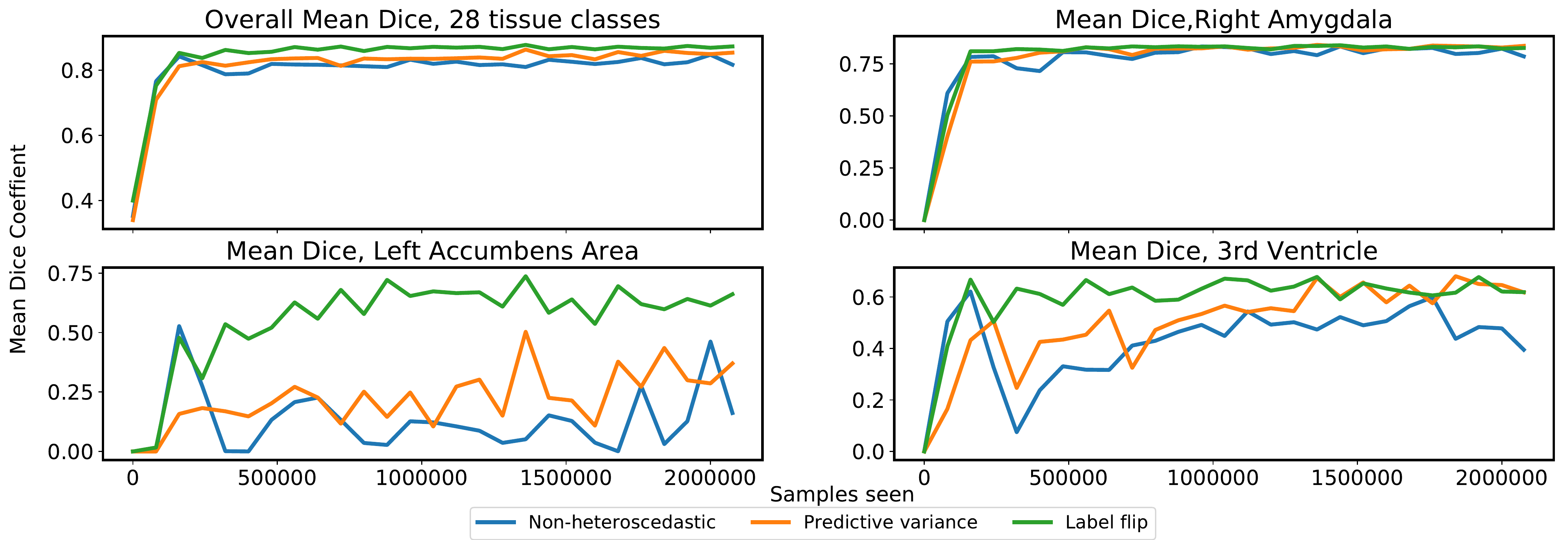}
    \caption{Learning curves (Mean Dice coefficient) over all compartments, and for selected individual tissue classes, on the six validation cases.}
    \label{fig:results}
\end{figure}

The plain network already reached its overall maximum mean Dice coefficient of (over all 28 tissue classes) after 160,000 samples were seen, after which overall performance declined and did not recover (See Figure~\ref{fig:results} and Table~\ref{tab:results}): this model achieved an overall mean Dice of $0.84\pm 0.12$ on the test set.
\begin{table}[]
    \centering
      \begin{filecontents*}{ABCD_selected_means.csv}
,Cortex,Putam.,Pallid.,Hipp.,Amyg.,Acc.,3rd Vent.,Overall
Predictive Variance,0.95,0.91,0.84,0.85,0.82,0.41,0.62,0.86
Label flip,0.95,0.91,0.85,0.87,0.82,0.66,0.60,0.88
Plain ($ \approx 2*{10^6}$ samples),0.95,0.91,0.84,0.86,0.78,0.17,0.37,0.83
Plain ($ \approx 2*10^5$ samples),0.95,0.89,0.80,0.82,0.76,0.60,0.65,0.85
\end{filecontents*}
\csvautotabular{ABCD_selected_means.csv}
      
          \_
    
    \caption{Mean Dice for selected compartments, for each of the three trained models, versus Freesurfer (n = 4096).  For lateral structures, results shown for left hemisphere only.  Results for the plain model shown at an early and a late epoch.}
    \label{tab:results}
\end{table}
By contrast, after training on approximately two million samples, neither heteroscedastic model showed signs of performance decline due to overfitting.  The model using predictive variance yielded a  mean Dice coefficient of $0.85 \pm 0.14$, and  the model using label-flip estimation yielded a  mean Dice coefficient of $0.87 \pm 0.11$.  Differences between overall Dice coefficients were significant ($p<10^{-6}$) On the level of individual structures, the plain network outperformed the predictive variance network on the 3rd ventricle and Accumbens areas, and outperformed the flip-probability network on the 3rd ventricle only.  On all other areas, the heteroscedastic models had better performance than the plain model.  All differences between models, on all compartments, were statistically significant ($p<10^{-6}$).

\section{Conclusion}
\begin{figure}
 \centering
    \includegraphics[width=12cm, keepaspectratio]{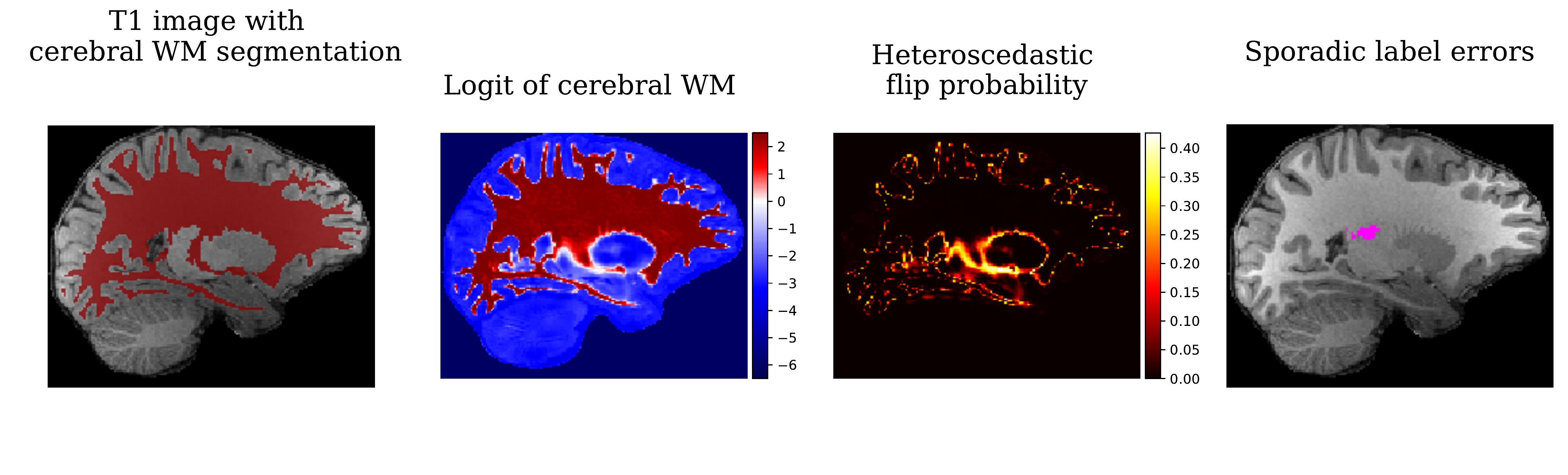}
    \caption {Example output of the label-flip heteroscedastic network, on the white-matter label in the left hemisphere: case taken from the ABCD cohort.}
    \label{fig:flip}
\end{figure}

Both heteroscedastic models outperformed the plain model over a range of brain segmentation tasks, with the model predicting label-flip probability achieving the best overall Dice coefficient.  Morevover, training of the heteroscedastic models was substantially more stable: as can be seen in Figure~\ref{fig:deepscan}, the performance of the plain model on difficult-to-segment structures such as the accumbens area and 3rd ventricle peaks early and then declines, before other regions such as the amygdala have reached optimal performance, while the label-flip model in particular is able to achieve good performance across all brain regions.  We expect that the label-flip maps, as seen in  Figure~\ref{fig:flip}, will be helpful for judging image/segmentation quality: however, the focus here is on performance in the small data, weak label domain. This is, we believe the first work in medical imaging to confirm the benefits of aleatoric uncertainty and its associated learned label smoothing for model performance.  Our proposed new notion of heteroscedastic network outperforms both the plain and the predictive variance network in this setting.  Further work is needed to verify if this benefit is still significant when training on hundreds or thousands of weakly labeled cases: in the other direction, preliminary results suggest that, for carefully selected training cases, it may even be possible to one-shot train a brain-segmentation network from a single training example. 

\bibliographystyle{splncs03}
\bibliography{MS}

\end{document}